\pdfoutput=1

\documentclass[11pt]{article}

\usepackage{acl}

\usepackage{times}
\usepackage{latexsym}

\usepackage[T1]{fontenc}

\usepackage[utf8]{inputenc}

\usepackage{microtype}

\usepackage{graphicx}
\usepackage{caption}
\usepackage{subcaption}
\usepackage{booktabs}
\usepackage{multirow}
\usepackage{xspace}
\usepackage{amsmath}
\usepackage{bm}
\usepackage{enumitem}
\usepackage{stfloats} 

\usepackage{xcolor}
\usepackage{soul}
\usepackage{makecell}
\newcommand{\esnli}{E-SNLI\xspace}
\newcommand{\fever}{FEVER\xspace}
\newcommand{\multirc}{MultiRC\xspace}

\newif\ifhidecomments
\hidecommentsfalse

\ifhidecomments
    \newcommand{\surya}[1]{}
    \newcommand{\sam}[1]{}
    \newcommand{\chenhao}[1]{}
\else
    \newcommand{\chenhao}[1]{\textcolor{blue}{[#1 ---\textsc{CT}]}}
    \newcommand{\sam}[1]{\textcolor{red}{[#1 ---\textsc{SC}]}}
    \newcommand{\surya}[1]{\textcolor{green}{[#1 ---\textsc{SK}]}}
\fi

\newcommand{\para}[1]{\noindent{\bf #1}\xspace}

\newcommand{\figref}[1]{Fig.~\ref{#1}}

\newcommand{\secref}[1]{\S\ref{#1}}

\newcommand{\learningfromexplanation}{learning from rationales\xspace}

\newcommand{\rationalizedinput}{rationalized input\xspace}
\newcommand{\rationalizedinputs}{rationalized inputs\xspace}
\newcommand{\vect}[1]{\ensuremath{\bm{#1}}}
\newcommand{\vecx}{\vect{x}}
\newcommand{\vecalpha}{\vect{\alpha}}

\title{What to Learn, and How: Toward Effective Learning from Rationales}

\author{Samuel Carton\\
  University of Chicago \\
  \texttt{carton@uchicago.edu   }
  \And
  Surya Kanoria\\
  University of Colorado Boulder \\
    \texttt{   surya.kanoria@colorado.edu   } \\

  \AND
  Chenhao Tan \\
  University of Chicago \\ 
  \texttt{   chenhao@uchicago.edu}
  }

\begin{document}
\maketitle
\begin{abstract}

Learning from rationales seeks to augment model prediction accuracy using human-annotated rationales (i.e. subsets of input tokens) that justify their chosen labels, often in the form of intermediate or multitask supervision. While intuitive, this idea has proven elusive in practice. 
We make two observations about human rationales via empirical analyses:
1) maximizing rationale supervision accuracy is not necessarily the optimal objective for improving model accuracy; 
2) human rationales vary in whether they provide sufficient information for the model to exploit for prediction.
Building on these insights, we propose several novel loss functions and learning strategies, and evaluate their effectiveness on three datasets with human rationales. Our results demonstrate consistent improvements over baselines in both label and rationale accuracy, including a 3\% accuracy improvement on MultiRC. Our work highlights the importance of understanding properties of human explanations and exploiting them accordingly in model training.

\end{abstract}

\begin{table}[!h]
\small
\centering

\begin{tabular}{@{}p{0.95\linewidth}@{}}
\toprule
\multicolumn{1}{c}{\textbf{(A) Unsupervised rationale}}                                                                                                                                                                                                                                                                                                                                                                                                                                                                                                                                                                                                                                                                                                                                                                                                                                                                                                                                                                                                                                                                           \\ \midrule
\definecolor{highlight}{RGB}{145, 224, 141}\sethlcolor{highlight}\hl{[CLS]}\hl{ }\hl{susan} wanted to have a \hl{birthday}\hl{ }\hl{party} . she called all of her \hl{friends} . she has five \hl{friends} . her \hl{mom} said that \hl{susan} can \hl{invite} them all to the \hl{party} . her \hl{first}\hl{ }\hl{friend}\hl{ }\hl{could} not \hl{go} to the \hl{party} because \hl{she} was \hl{sick} . \hl{her}\hl{ }\hl{second}\hl{ }\hl{friend} was going out \hl{of} town . her \hl{third}\hl{ }\hl{friend} was not so sure if her \hl{parents} would let her . the \hl{fourth}\hl{ }\hl{friend} said \hl{maybe} . the fifth \hl{friend} could go to the \hl{party} for sure \hl{.}\hl{ }\hl{susan} was a little \hl{sad}\hl{ }\hl{.} on the day of the \hl{party} , all \hl{five}\hl{ }\hl{friends} showed up . each \hl{friend} had a \hl{present} for \hl{susan} . \hl{susan} was happy and sent each \hl{friend} a thank you card the next week . \hl{[SEP]}\hl{ }\hl{how}\hl{ }\hl{many}\hl{ }\hl{people}\hl{ }\hl{did}\hl{ }\hl{susan}\hl{ }\hl{call}\hl{ }\hl{?}\hl{ }\hl{|}\hl{ }\hl{|}\hl{ }\hl{5}\hl{ }\hl{[SEP]} \\ \midrule
{\textbf{Prediction: \color[HTML]{FE0000} False}}                                                                                                                                                                                                                                                                                                                                                                                                                                                                                                                                                                                                                                                                                                                                                                                                                                                                                                                                                                                                                                                                                   \\ \midrule
\multicolumn{1}{c}{\textbf{(B) Human rationale}}                                                                                                                                                                                                                                                                                                                                                                                                                                                                                                                                                                                                                                                                                                                                                                                                                                                                                                                                                                                                                                                                                        \\ \midrule
\definecolor{highlight}{RGB}{250, 215, 160 }\sethlcolor{highlight}\hl{[CLS]} susan wanted to have a birthday party . \hl{she}\hl{ }\hl{called}\hl{ }\hl{all}\hl{ }\hl{of}\hl{ }\hl{her}\hl{ }\hl{friends}\hl{ }\hl{.}\hl{ }\hl{she}\hl{ }\hl{has}\hl{ }\hl{five}\hl{ }\hl{friends}\hl{ }\hl{.} her mom said that susan can invite them all to the party . her first friend could not go to the party because she was sick . her second friend was going out of town . her third friend was not so sure if her parents would let her . the fourth friend said maybe . the fifth friend could go to the party for sure . susan was a little sad . on the day of the party , all five friends showed up . each friend had a present for susan . susan was happy and sent each friend a thank you card the next week . \hl{[SEP]}\hl{ }\hl{how}\hl{ }\hl{many}\hl{ }\hl{people}\hl{ }\hl{did}\hl{ }\hl{susan}\hl{ }\hl{call}\hl{ }\hl{?}\hl{ }\hl{|}\hl{ }\hl{|}\hl{ }\hl{5}\hl{ }\hl{[SEP]}                                                                                                                                            \\ \midrule
\textbf{Prediction: True}                                                                                                                                                                                                                                                                                                                                                                                                                                                                                                                                                                                                                                                                                                                                                                                                                                                                                                                                                                                                                                                                                                           \\ \bottomrule
\end{tabular}

\caption{An example of unsupervised versus human-provided rationale in MultiRC. The unsupervised model struggles to localize its attention and makes an incorrect prediction. The same model makes a correct prediction by only looking at the human rationale.}
\label{tab:intro_rationale_example}
\end{table}

\section{Introduction}

In the past several years, explainability has become a prominent issue in machine learning, addressing concerns about the safety and ethics of using large, opaque models for decision-making. As interest has grown in explanations for understanding model behavior, so has interest grown in soliciting gold-standard explanations from human annotators and using them to inject useful inductive biases into models \cite{hase_when_2021}. Many such explanation datasets have become available recently \cite{wiegreffe_teach_2021}.

A common format for explanations in NLP is the \textit{rationale}, 
a subset of input tokens that are relevant to the decision.
A popular architecture for generating such explanations is the \textit{rationale model}, an explain-then-predict architecture which first 
extracts a rationale from the input and
then makes a prediction from 
the rationale-masked text (that is, only the tokens included in rationale) \cite{lei_rationalizing_2016, deyoung_eraser_2019}. Without  external supervision on this rationale, we typically pursue parsimony via a sparsity objective.
Table~\ref{tab:intro_rationale_example}A shows an example unsupervised rationale.

With the benefit of a human-annotated rationale for the true label, we can begin to understand model mistakes in terms of reliance on inappropriate features (and correct them). In the example above, the unsupervised rationale suggests that the model's mistake is due to missing key information about how many friends Susan has (i.e., ``five''). Forcing the model to see these key tokens by only using the human rationale as the input 
fixes this mistake (Table \ref{tab:intro_rationale_example}B). 
Prior work has shown that this is not a fluke. For some datasets, human rationales consistently improve model accuracy over baseline when used as an input mask, by orienting model attention toward informative tokens and away from confounding ones
\cite{carton_evaluating_2020}.

Knowing that human rationales contain useful predictive signal, the key question becomes: \textbf{can we improve model prediction accuracy by incorporating human rationales into training?} 

Numerous approaches to using human rationales in training have been tried, including: regularizing the parameters of a (linear) model \cite{zaidan_using_2007}; regularizing model output gradients \citep{ross_right_2017}; regularizing internal transformer attention weights \cite{jayaram_human_2021}; and direct supervision on a rationale model \cite{deyoung_eraser_2019}, which serves as our baseline approach in this paper. These approaches have generally failed to significantly improve model prediction accuracy \cite{hase_when_2021}. 

A quality these prior approaches have in common is treating human rationales as \textit{internally and collectively uniform} in predictive utility. That is, any token included in the human rationale is treated as equally important to include in the input representation; vice versa for tokens excluded. Furthermore, all human rationales are weighted equally.

The reality, we demonstrate empirically via ablation studies in \secref{sec:analysis}, is that the predictive utility of human rationales is distributed unevenly between tokens in a rationale, and unevenly between rationales in a dataset. Based on this analysis, we suggest that learning objectives which weight every token equally (accuracy in the case of direct supervision), and every rationale equally, are not optimal for improving downstream model accuracy. 

We operationalize these hypotheses in four distinct modifications to the baseline rationale model architecture. Three of these modify the naive token-wise accuracy supervision objective, and the fourth implements ``selective supervision'', ignoring unhelpful human rationales in training.

Evaluating on three datasets, our proposed methods produce varying levels of improvement over both a baseline BERT model and a baseline BERT-to-BERT supervised rationale model,
ranging from  substantial for \multirc (3\%) to marginal for \esnli (0.4\%).
Additionally,
 our methods also improve rationale prediction performance.

Taken together, our results demonstrate the importance of considering the variance of predictive utility both between and within human rationales as a source of additional training signal.
Our proposed modifications 
help pave the way toward truly effective and general \learningfromexplanation.

\section{Related Work}

\subsection{Rationalization}

The extractor-predictor rationale model proposed by \citet{lei_rationalizing_2016} and described in more detail in \secref{sec:methods}, is an approach to feature attribution, which is one among many families of explanation methods (see \citet{vilone_explainable_2020} for a recent survey).

Recent work has extended the original architecture in various ways, including replacing the use of reinforcement learning with differentiable binary variables \cite{bastings_interpretable_2020,deyoung_eraser_2019}, alternatives to the original sparsity objective \cite{paranjape_information_2020, antognini_rationalization_2021}, and additional modules which change the interaction dynamics between the extractor and predictor \cite{carton_extractive_2018, yu_rethinking_2019, chang_invariant_2020}. Pipeline models \cite{lehman_inferring_2019} are similar, but train the two modules separately rather than end-to-end. 

Rationale models are a powerful approach to NLP explanations because of how specific objectives can be put on the properties of the rationale, but they have some downsides. First, they are unstable, the extractor often collapsing to all-0 or all-1 output \cite{deyoung_eraser_2019, yu_rethinking_2019}. We introduce an engineering trick in \secref{sec:methods} that appears to lessen this risk. Also, with end-to-end training comes the risk of information leakage between the extractor and predictor \cite{jethani_have_2021, hase_leakage-adjusted_2020, yu_understanding_2021}. This idea of leakage plays a part in how we estimate explanation predictive utility in section \secref{sec:analysis}.

\subsection{Learning from Explanations}

\citet{wiegreffe_teach_2021} present a review of explainable NLP datasets, a number of which have been incorporated into the ERASER collection and benchmark \citep{deyoung_eraser_2019}. 

Early work in learning from human explanations include \citet{zaidan_using_2007} and \citet{druck_active_2009}, and a line of work termed ``explanatory debugging'' \citep{kulesza_principles_2015,lertvittayakumjorn_explanation-based_2021}. More recent work spans a variety of approaches, categorized by \citet{hase_when_2021} into regularization (e.g., \citet{ross_right_2017}), data augmentation (e.g., \citet{hancock2018training}), and supervision over intermediate outputs (e.g., \citet{deyoung_eraser_2019, jayaram_human_2021}).

Significant improvements to model accuracy as a result of explanation learning have proven elusive. Studies occasionally claim such improvement, such as \citet{rieger_interpretations_2020}, which observes general improvements on a medical vision task.
More commonly their claims pertain to secondary objective such as explanation quality (e.g., \citet{plumb_regularizing_2020}), 
robustness (e.g., \citet{ross_right_2017}, \citet{srivastava_robustness_2020}),
or few-shot learning (e.g., \citet{yao_refining_2021}). 
\citet{hase_when_2021} gives an overview of the problem and discusses circumstances under which learning from explanations is liable to work. Our paper contributes to this discussion by considering the variance of training signal quality both within and between human rationales, and how to exploit these variances.

\section{Data}

We consider three datasets in this work.
All three are document-query text comprehension tasks, where the task is to determine whether the query is true or false given the document.
We use the train, development, test splits offered by \citet{deyoung_eraser_2019}.
Table~\ref{tab:data} shows the basic statistics of each dataset based on the training set.

\begin{itemize}[leftmargin=*, topsep=-2pt, itemsep=-2pt]

\item \textbf{\multirc} \citep{khashabi2018looking}. A reading comprehension dataset of 32,091 document-question-answer triplets that are true or false.
Rationales consist of 2-4 sentences from a document that are required to answer the given question.

\item \textbf{\fever} \cite{thorne2018fever}. A fact verification dataset of 76,051 snippets of Wikipedia articles paired with 
claims that they support or refute. 
Rationales consist of a single contiguous sub-snippet, so the basic unit of rationale is sentence.

\item \textbf{\esnli} \cite{camburu2018snli}. A textual entailment dataset of 568,939 short snippets and claims for which each snippet either refutes, supports, or is neutral toward. 
Input texts are much shorter than \multirc and \fever, and rationales are at the token level.

\end{itemize}

\begin{table}[t]
    \small
    \centering
    \begin{tabular}{l>{\raggedleft\arraybackslash}p{1.5cm}>{\raggedleft\arraybackslash}p{1.5cm}>{\raggedleft\arraybackslash}p{1.5cm}}
    \toprule
        Dataset & {Text length} & { Rationale length} & Rationale granularity \\
        \midrule
        \multirc & 336.0 & 52.0 & sentence \\
        \fever & 355.9 & 47.0 & sentence \\
        \esnli & 23.5 & 6.1 & token \\
         \bottomrule
    \end{tabular}
    \caption{Basic statistics of the datasets.} 
    \label{tab:data}
\end{table}

\begin{figure*}[h]
    \centering
    \begin{subfigure}[t]{0.32\textwidth}
        \centering
        \includegraphics[width=0.9\textwidth, trim=0 0 230 0, clip]{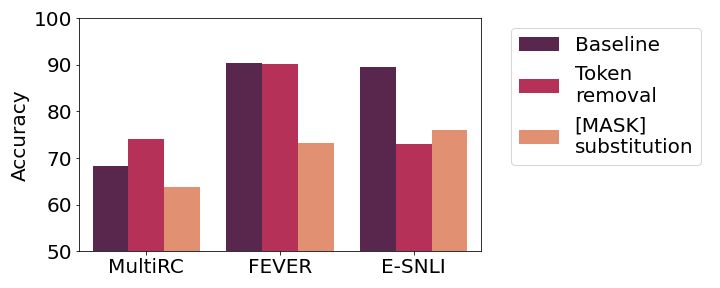}
        \caption{Fine-tuned on full input (unadapted).}
        \label{fig:combined_unadapted_sufficiency_accuracy}
    \end{subfigure}
    \begin{subfigure}[t]{0.6\textwidth}
        \centering
        \includegraphics[width=0.7\textwidth]{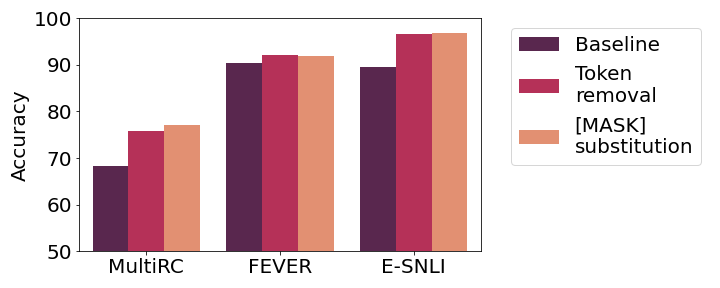}
        \caption{Fine-tuned on both full and human-rationalized input (adapted).}
        \label{fig:combined_adapted_sufficiency_accuracy}
    \end{subfigure}
   
    \caption{Baseline performance vs. human sufficiency-accuracy for \rationalizedinputs with token removal and [MASK] token substitution.
    As \rationalizedinputs are different from the full text inputs that the original training set includes, we build a calibrated model where the model is trained on both full text inputs and \rationalizedinputs. }
    \label{fig:combined_sufficiency_accuracy}
\end{figure*}

\section{Analysis}
\label{sec:analysis}

To understand properties of human rationales for the purpose of \learningfromexplanation,
we analyze the effect of human rationales when they are used 
as inputs to a trained model.

\subsection{Human Rationales have Predictive Utility}

A basic question about the viability of \learningfromexplanation~is whether human rationales bear the potential for 
improving model performance. That is, do human explanations successfully reveal useful tokens while occluding confounding tokens, such that a model evaluated only on the revealed tokens is able to get improved performance relative to the full input? We refer to such rationale-redacted inputs as {\em \rationalizedinputs}.

\def \dataset {multirc}
\def \basedir {new_plots}
\begin{figure*}[h]
\centering
\begin{subfigure}[t]{0.41\textwidth}
    \centering
    \includegraphics[width=0.9\textwidth]{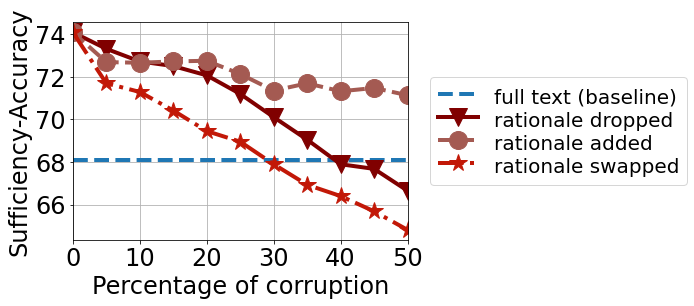}
    \caption{All samples}
    \label{fig:multirc_perturbation_plots_all}
\end{subfigure}
\begin{subfigure}[t]{0.28\textwidth}
    \centering
    \includegraphics[width=0.85\textwidth, trim=0 0 270 0, clip]{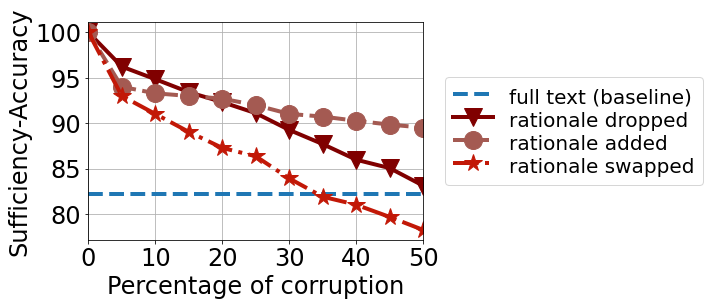}
    \caption{Human sufficiency-accuracy = 1}
    \label{fig:multirc_perturbation_plots_suffacc_1}
\end{subfigure}
\begin{subfigure}[t]{0.28\textwidth}
    \centering
    \includegraphics[width=0.83\textwidth, trim=0 0 270 0, clip]{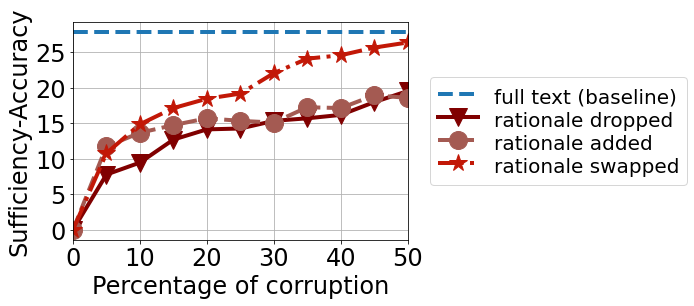}
    \caption{Human sufficiency-accuracy = 0}
    \label{fig:multirc_perturbation_plots_suffacc_0}
\end{subfigure}
\caption{Sufficiency-accuracy of human rationales on baseline BERT model with increasing levels of corruption via swaps, drops and additions.
Model performance decreases quickly when we drop rationale tokens
, but stays high as we add non-rationale tokens.
These effects are moderated by HSA.}
\label{fig:multirc_perturbation_plots}
\end{figure*}

We define \textit{sufficiency-accuracy} (\textbf{SA}) as how accurate the model is across a corpus of \rationalizedinput.
This is an aggregate measure, similar to \textit{sufficiency} as defined in \citet{deyoung_eraser_2019} but focused on absolute performance rather than similarity to baseline model output.
We refer to the sufficiency-accuracy of the human rationales as \textit{human sufficiency-accuracy} (\textbf{HSA}).

Estimating sufficiency-accuracy is problematic. The natural way to probe whether the tokens in a rationale are sufficient for an accurate prediction is to remove the non-included tokens from the input, run the model on just the included tokens, and assess its accuracy. But a version of the input where a majority of tokens are removed or masked (by a [MASK] special token in the case of BERT), is out-of-distribution relative to the training data, which has no removal or masking. This difference may lead to unpredictable output from the model when tested on masked input. This \textbf{masking-is-OOD} problem has not received much discussion in the literature, though \citet{jacovi_aligning_2021} propose to mitigate it with random masking during model training. The effect of this problem will be to underestimate the sufficiency-accuracy of rationales tested against an un-adapted model. 

The opposite problem stems from overfitting rather than OOD issues: \textbf{label leakage}. A human rationale may contain signal about the true label that goes beyond the semantics of the tokens included in the rationale, and a model trained on human-rationalized input may learn to pick up on these spurious signals. A known example is in \esnli, where annotators had different explanation instructions based on their chosen label. This issue is discussed in several recent papers \cite{yu_understanding_2021, jethani_have_2021, hase_leakage-adjusted_2020}, albeit mostly concerning model-generated rather than human explanations. The effect of this problem will be to overestimate the sufficiency-accuracy of rationales tested against an adapted model. 

\figref{fig:combined_sufficiency_accuracy} shows sufficiency-accuracy results for human rationales on both unadapted and adapted models. We expand on the analysis presented by \citet{carton_evaluating_2020} by showing results for both masking-via-removal and masking-via-[MASK]-token-substitution.

\figref{fig:combined_unadapted_sufficiency_accuracy} shows that token removal suffers less from the masking-is-OOD problem on an unadapted model than [MASK] token substitution. [MASK] token substitution results in lower accuracy across the board, while removal improves baseline accuracy for \multirc, matches it for \fever, and lowers it for \esnli. 

With adaptation (\figref{fig:combined_adapted_sufficiency_accuracy}), token removal and [MASK] token substitution have near-identical effects, improving accuracy by a large margin for \multirc and \esnli, and a small margin for \fever. The near-100\% sufficiency-accuracy for \esnli is probably due to label leakage.

If an unadapted model is liable to underestimate sufficiency model, and an adapted model to overestimate, then we suggest that the potential benefit of \learningfromexplanation lies somewhere between the two. On this hypothesis, this figure suggests that \multirc has a large potential benefit, \fever a small one, and \esnli an unclear benefit depending on how much of the predictive utility of \esnli rationales is due to label leakage. The results in \secref{sec:p06_results} ultimately bear out these expectations.

\subsection{Importance of Rationale Accuracy}
\label{subsec:rationale-recall-analysis}

We focus on \multirc, where evaluating a non-rationale-adapted fine-tuned BERT model on human-rationalized data results in a sufficiency-accuracy of 74\%, a significant improvement over the normal test accuracy of 68\%. But how robust is this improvement to rationale prediction error? We examine how the sufficiency-accuracy of human rationales changes as they are corrupted by random addition, dropping, and swapping of tokens.

In this analysis, an $N\%$ drop removes $N\%$ of tokens from each rationale in the dataset, reducing recall to $100-N$. An $N\%$ addition adds tokens numbering $N\%$ the size of each rationale, from the set of non-rationale tokens, reducing precision to $\frac{100}{100+N}$. An $N\%$ swap performs both operations, swapping $N\%$ of rationale tokens for the same number of non-rationale tokens.

The ``dropped'' curve in \figref{fig:multirc_perturbation_plots_all}
shows that human rationales afford improved accuracy over the baseline until roughly 40\% of tokens have been dropped from them, suggesting that a minimum of 60\% recall is needed to derive an advantage from human rationales over the full input. Per the ``added'' curve, adding the same number of irrelevant tokens to the rationale has a much less severe impact on accuracy, suggesting 
that errors of omission are significantly worse than errors of inclusion for \learningfromexplanation.

\figref{fig:multirc_perturbation_plots_suffacc_1} and \ref{fig:multirc_perturbation_plots_suffacc_0} respectively show the effect of this perturbation on high- and low-sufficiency-accuracy human rationales, which constitute 74\% and 26\% of rationales respectively for this model. High-SA rationales follow a similar trend to the whole population, but the recall requirement is lower than \figref{fig:multirc_perturbation_plots_all} to exceed model accuracy with the full input (the ``dropped'' curve meets the blue line at 50\%).
In comparison, low-SA rationales demonstrate interesting properties. 
These rationales actually have a sabotaging effect in a quarter of cases: the model would have an accuracy of 27\% with the full input, which is lowered to 0\% by the presence of these rationales. Also, addition and dropping have a similar effect in mitigating this sabotage.
Similar results hold on \fever and \esnli except the apparent required recall is much higher ($>$90\%) for both methods (see the appendix), indicating challenges for \learningfromexplanation on these datasets.

In summary, our analyses inspire two general observations about \learningfromexplanation:
1) moving away from naive accuracy (toward recall, for example) as a rationale supervision objective, and 2) focusing on useful rationales over harmful ones.

\section{Methods}
\label{sec:methods}

We propose architecture changes based on these insights.
Our code is available at {\small\url{https://github.com/ChicagoHAI/learning-from-rationales}}.

\subsection{Background and Baseline Models}

Our training data include input tokens, their corresponding rationales, and labels.
Formally, an instance is denoted as $(\vecx, \vecalpha, y)$, where $\vecx = (\vecx_1, \ldots, \vecx_L)$ is a text sequence of length $L$ and human rationale $\vecalpha$ of the same length.
$\vecalpha_i=1$ indicates that token $\vecx_i$ is part of the rationale (and relevant for the prediction), $\vecalpha_i=0$ otherwise.

We use HuggingFace's BERT-base-uncased \citep{devlin_bert:_2018,wolf_huggingfaces_2020} as the basis for our experiments and analysis. Used in the standard way, BERT ignores $\vecalpha$ and is fine-tuned on tuples of $(\vecx, y)$. This is our simplest baseline.

\begin{figure}
    \centering
    \includegraphics[width=0.95\columnwidth]{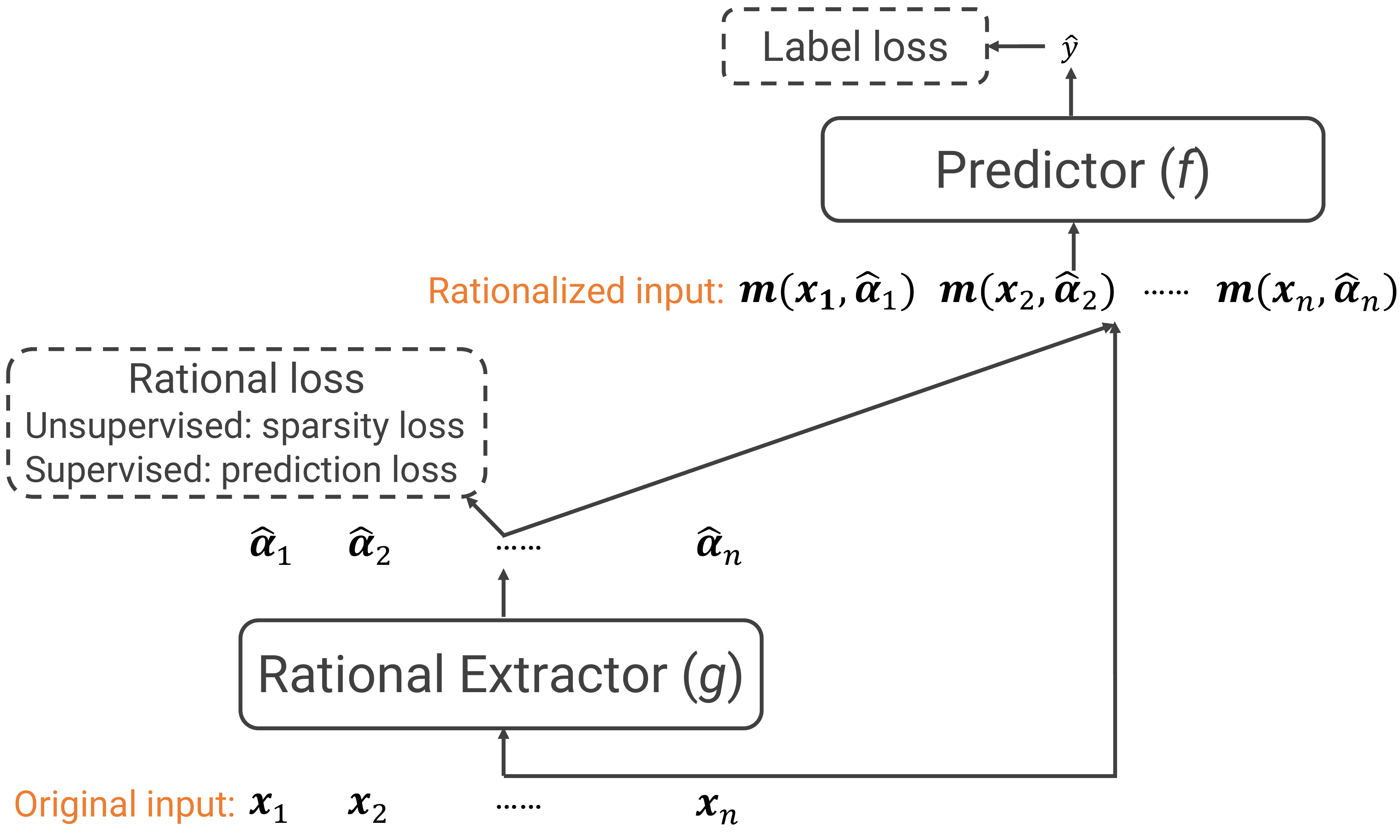}
    \caption{Illustration of our multi-task framework. Our main innovation lies in how we define rationale loss for the supervised case and the masking function $m$.}
    \label{fig:model}
\end{figure}

\para{Rationale model.}
We use the rationale model of \citet{lei_rationalizing_2016} for both supervised and unsupervised rationale generation, in its updated BERT-to-BERT form \cite{deyoung_eraser_2019}. 
This model consists of two 
BERT modules: a rationale extractor $g$ that generates a binary attention mask $\hat{\vecalpha}$ as the rationale, and a predictor $f$ which makes a prediction using the \rationalizedinput via a masking function $m$ on $\vecx$ and $\hat{\vecalpha}$ (\figref{fig:model}):
\begingroup\abovedisplayskip=4pt \belowdisplayskip=4pt
\begin{displaymath}
\begin{aligned}
&g(\vecx) \rightarrow \hat{\vecalpha}, \\
&f(m(\vecx, \hat{\vecalpha})) \rightarrow \hat{y}.
\end{aligned}
\end{displaymath}
\endgroup

The two components are trained in tandem. In the unsupervised scenario, the joint objective function consists of a prediction loss term and a rationale sparsity term, encouraging the model to retain only those tokens in $\vecx$ that are necessary for accurate prediction:
\begingroup\abovedisplayskip=4pt \belowdisplayskip=4pt
$$\mathcal{L}_u=\mathcal{L}_p(y,\hat{y}) + \lambda_{sp}||\hat{\vecalpha}||,$$
\endgroup
where $\mathcal{L}_p$ is typically cross entropy. 

In the supervised scenario, given a human rationale $\vecalpha$, we replace the sparsity objective with a rationale supervision objective:
\begingroup\abovedisplayskip=4pt \belowdisplayskip=4pt
$$\mathcal{L}_{su}=\mathcal{L}_p(y,\hat{y}) + \frac{\lambda_{su}}{L}\sum_{i=1}^L \mathcal{L}_p(\vecalpha_i, \hat{\vecalpha}_i),$$
\endgroup
where $\lambda_{su}$ is a hyperparameter that controls the weight of rationale loss compared to label loss.

Each of these scenarios represents a baseline for our experiment. 
We refer to the unsupervised version as {\em unsupervised rationale model}, and the supervised version as {\em supervised rationale model}.

\para{Implementation details.}
The original \citet{lei_rationalizing_2016} model generates binary rationales by 
Bernoulli sampling from continuous probability values produced by the generator, 
and uses the REINFORCE algorithm \citep{williams_simple_1992} to propagate approximate gradients through this non-differentiable operation. 

We instead use Gumbel Softmax \citep{jang_categorical_2017} to generate differentiable approximate binary rationale masks. In this framework,
the generator produces logits $\vect{z_i}$ 
to which are added random noise $G \sim Gumbel(0,1)$, before applying a softmax
to produce 
class probabilities $\vect{c_i}$. This approximates 
a discrete distribution parameterized by $e^{\vect{z_i}}$. We then use the positive class probability $\vect{c}_i^1$ as the rationale value $\hat{\vecalpha}_i$.
\begingroup\abovedisplayskip=4pt \belowdisplayskip=4pt
$$\vect{c}_i = \operatorname{softmax}(\vect{z_i}+\vect{G} \sim Gumbel(0,1)); \hat{\vecalpha}_i = \vect{c}_i^1 $$
\endgroup

\para{Generating stable rationales.}
We find it helpful as an engineering trick to pre-train the predictor layer of this model on the full input before co-training the predictor and extractor on the joint objective. This step appears to mitigate some of the issues this model has with rationale collapse, noted for example by \citet{deyoung_eraser_2019}. 

Given $\hat{\vecalpha}_i$,
we mask non-rationale tokens by multiplicatively substituting the [MASK] token vector across their vector representations, analogously to what is done during the MASK-LM pretraining of the BERT model:
\begingroup\abovedisplayskip=4pt \belowdisplayskip=4pt
$$m_s(\vecx_i, \hat{\vecalpha}_i) = \hat{\vecalpha}_i \cdot \vect{e}_i + (1- \hat{\vecalpha}_i) \cdot \vect{e}_{\text{[MASK]}},$$
\endgroup
where $\vect{e}_i$ represents the embedding associated with $\vecx_i$ and $\vect{e}_{\text{[MASK]}}$ is the embedding for the [MASK] token.
We never mask special tokens [CLS] or [SEP], and we set $\hat{\vecalpha}_i=1$ for the query in \multirc and \fever as well because the query is always part of human rationales in these two datasets.

\subsection{Learning from Human Rationales}

Inspired by the analysis in \secref{sec:analysis}, we propose four strategies for improving the efficacy of \learningfromexplanation: 1) tuning class weights for rationale supervision; 2) enforcing sentence-level rationalization; 3) using non-occluding ``importance embeddings''; 
and 4) selectively supervising only rationales with high sufficiency-accuracy. 
The first three are designed to loosen the supervision's dependence on flat tokenwise accuracy, while the last tries to operationalize our observations about helpful versus non-helpful rationales.

\para{Class weights.}
Rationales may become more effective enablers of model prediction accuracy at different balances of precision and recall. We can adjust this balance simply by using differing weights to positive and negative classes in rationale supervision:
\begingroup\abovedisplayskip=4pt \belowdisplayskip=4pt
$$\mathcal{L}_w=\mathcal{L}_p(y,\hat{y}) + \frac{1}{L}\sum_{i=1}^L (1 + \lambda_{su}^1\alpha_i) \mathcal{L}_p(\alpha_i, \hat{\alpha}_i),$$
\endgroup
where $\lambda_{su}^1$ controls the relative weight of rationale vs. non-rationale tokens.
In particular, as we will discuss in \secref{sec:analysis}, we find that 
increased recall is associated with increased model accuracy. Thus, we explore several values for $\lambda_{su}^1$ in our experiment to encourage higher recall.

\para{Sentence-level rationalization.}
Another divergence from strict token-wise accuracy is to rationalize at the sentence rather than the token level. 
Given a function $sent$ mapping a token $x_i$ to its corresponding sentence $s$ consisting of tokens $\{...,x_i,...\}$, 
we average token-level logits $\vect{z}_i$ across each sentence to produce a binary mask at the sentence level and then propagate that mask value to all sentence tokens:
\begingroup\abovedisplayskip=4pt \belowdisplayskip=4pt
$$\hat{\vect{\alpha}}_i = \hat{\vect{\alpha}}^{s}_{sent(i)},$$
\endgroup
where $\vect{z}^s = \frac{1}{|\{i | sent(i) = s\}|}\sum_{\{i | sent(i) = s\}}{\vect{z}_i}$ is used to generate $\hat{\vect{\alpha}}^{s}_{sent(i)}$.

\para{Importance embeddings.}
Another way to mitigate the impact of false negatives in predicted rationales is for these negatives to still remain visible to the predictor. This variant uses additive embeddings for rationalization rather than occluding masks,
using a two-element embedding layer $\vect{e}$ constituting one embedding for rationale tokens and one for nonrationale tokens, added to the input vectors according to the predicted rationale. This way, input tokens are tagged as important or unimportant, but the predictor $f$ has the freedom to learn how to engage with these tags for maximum label accuracy, rather than being fully blinded to ``unimportant'' tokens. 
\begingroup\abovedisplayskip=4pt \belowdisplayskip=4pt
$${\scriptstyle m_e(\vecx_i,\hat{\vecalpha}_i) = \vect{e}_i + (1-\hat{\vecalpha}_i) \cdot \vect{e}_{\operatorname{non-rationale}} + \hat{\vecalpha}_i\cdot \vect{e}_{\operatorname{rationale}}}.$$
\endgroup
An important drawback of this approach is that the predictor now has access to the full input instead of only the \rationalizedinput, so these rationales provide a weak guarantee that important tokens are actually used to make predictions. This method also represents a large distribution shift from full text, so we find it necessary to calibrate the predictor using human rationales, as described in \figref{fig:combined_adapted_sufficiency_accuracy}. 

\para{Selective supervision.}
Our fourth modification attempts to improve rationale prediction performance on high-sufficiency-accuracy rationales by selectively supervising only on human rationales with this property, ignoring those where human rationales do not allow a correct prediction. 

Specifically, for every training batch, we use the true human rationales $\vecalpha$ as an input mask for the BERT predictor to get the HSA for each document.
HSA then serves as a weight on the human rationale supervision during the main training batch: 
\begingroup\abovedisplayskip=4pt \belowdisplayskip=4pt
$${\scriptstyle \mathcal{L}_{ss}=\mathcal{L}_p(y,\hat{y}) + I(y = f(m(\vecx, \vecalpha)))\frac{\lambda_{su}}{L}\sum_{i=1}^L \mathcal{L}_p(\vecalpha_i, \hat{\vecalpha}_i)}.$$
\endgroup

By weighting supervision this way, we hope to ignore 
low-quality
human rationales during training and focus instead on those that enable good accuracy.

\section{Results}
\label{sec:p06_results}

\begin{table*}[]
\scriptsize
\centering
\setlength\tabcolsep{5pt}
\begin{tabular}{@{}llccccccccc@{}}
\toprule
\multirow{2}{*}{\textbf{Dataset}} & \multicolumn{1}{c}{\multirow{2}{*}{\textbf{Model}}} & \multirow{2}{*}{\textbf{Acc.}} & \multicolumn{3}{c}{\textbf{Rationale prediction}}                     & \multirow{2}{*}{\textbf{\begin{tabular}[c]{@{}r@{}}Human \\ Suff. Acc.\end{tabular}}} & \multicolumn{4}{c}{\textbf{Methods}}                                                                                                                                                                                                \\ \cmidrule(lr){4-6} \cmidrule(l){8-11} 
                                  & \multicolumn{1}{c}{}                                &                                & \textbf{F1}           & \textbf{Prec.}        & \textbf{Rec.}         &                                                                                       & \multicolumn{1}{c}{\textbf{Masking}} & \multicolumn{1}{c}{\textbf{Granularity}} & \textbf{\begin{tabular}[c]{@{}c@{}}Pos. class \\ weight\end{tabular}} & \textbf{\begin{tabular}[c]{@{}c@{}}Selective \\ supervision\end{tabular}} \\ \midrule
\multirow{4}{*}{\textbf{MultiRC}} & \textbf{BERT baseline}                              & 68.1                           & \multicolumn{1}{c}{-} & \multicolumn{1}{c}{-} & \multicolumn{1}{c}{-} & 73.9                                                                                  & \multicolumn{1}{c}{-}                & Tokens                                   & -                                                                     & -                                                                         \\
                                  & \textbf{Unsupervised rationale model}                      & 67.2                           & 22.2                  & 18.5                  & 27.9                  & 71.2                                                                                  & [MASK]                               & Tokens                                   & -                                                                     & -                                                                         \\
                                  & \textbf{Supervised rationale model}                        & 67.0                           & 46.5                  & 41.5                  & 52.9                  & 70.8                                                                                  & [MASK]                               & Tokens                                   & \multicolumn{1}{r}{1.0}                                               & \multicolumn{1}{r}{No}                                                    \\
                                  & \textbf{Best overall model}                         & \textbf{71.2}                  & \textbf{57.1}                  & \textbf{44.9}                  & \textbf{78.4}                  & \textbf{74.5}                                                                                  & Embeddings                           & Sentences                                & \multicolumn{1}{r}{5.0}                                               & \multicolumn{1}{r}{No}                                                    \\ \midrule
\multirow{4}{*}{\textbf{FEVER}}   & \textbf{BERT baseline}                              & 90.2                           & \multicolumn{1}{c}{-} & \multicolumn{1}{c}{-} & \multicolumn{1}{c}{-} & 89.4                                                                                  & \multicolumn{1}{c}{-}                & Tokens                                   & -                                                                     & -                                                                         \\
                                  & \textbf{Unsupervised rationale model}                      & 88.3                           & 22.6                  & 20.5                  & 25.1                  & 88.7                                                                                  & [MASK]                               & Tokens                                   & -                                                                     & -                                                                         \\
                                  & \textbf{Supervised rationale model}                        & 90.7                           & 68.4                  & 61.7                  & 76.7                  & 91.1                                                                                  & [MASK]                               & Tokens                                   & \multicolumn{1}{r}{1.0}                                               & \multicolumn{1}{r}{No}                                                    \\
                                  & \textbf{Best overall model}                         & \textbf{91.5}                  & \textbf{81.2}                  & \textbf{83.5}                  & \textbf{79.1}                  & \textbf{91.6}                                                                                  & Embeddings                           & Sentences                                & \multicolumn{1}{r}{1.0}                                               & \multicolumn{1}{r}{No}                                                    \\ \midrule
\multirow{4}{*}{\textbf{E-SNLI}}  & \textbf{BERT baseline}                              & 89.7                           & \multicolumn{1}{c}{-} & \multicolumn{1}{c}{-} & \multicolumn{1}{c}{-} & 73.9                                                                                  & \multicolumn{1}{c}{-}                & Tokens                                   & -                                                                     & -                                                                         \\
                                  & \textbf{Unsupervised rationale model}                      & 88.9                           & 40.6                  & 28.2                  & 72.6                  & 85.0                                                                                  & [MASK]                               & Tokens                                   & -                                                                     & -                                                                         \\
                                  & \textbf{Supervised rationale model}                        & 87.8                           & 58.7                  & \textbf{47.7}                  & 76.0                  & 89.4                                                                                  & [MASK]                               & Tokens                                   & \multicolumn{1}{r}{1.0}                                               & \multicolumn{1}{r}{No}                                                    \\
                                  & \textbf{Best overall model}                         & \textbf{90.1}                           & \textbf{59.6}                  & 45.5                  & \textbf{86.2}                  & \textbf{92.3}                                                                                  & Embeddings                           & Tokens                                   & \multicolumn{1}{r}{3.0}                                               & \multicolumn{1}{r}{No}                                                    \\ \bottomrule
\end{tabular}
\caption{Best-performing model variant compared to baseline models. 
}
\label{tab:baseline_results}
\end{table*}

\subsection{Experiment Setup}

Our goal in this experiment is to understand the impact of 
our four proposed model/training modifications.
We do this with a comprehensive scan: 
We try three positive rationale supervision class weights $\lambda_{su}^1$ (\{0, 2, 4\}), and toggle sentence-level rationalization, importance embedding, selective supervision on and off.
In addition, we vary rationale supervision loss weight $\lambda_{su}$ in \{0.5, 1, 2\}.
This resulted in 72 models for \multirc and \fever, and 36 models for \esnli (for which sentence-level rationalization is not applicable). 

The best resultant model is our {\em best overall model}.
The best model with $\lambda_{su^1}=1$ (i.e., identical class weights for human rationales) and no other learning strategy enabled is our baseline {\em supervised rationale model}.
We additionally train three {\em unsupervised rationale models} with sparsity weights 0.15, 0.25, and 0.35, selecting as representative the one which produced the sparsest rationales while maintaining a reasonable level of accuracy (because in this architecture, there is invariably a trade-off between accuracy and sparsity).

To evaluate the performance of our models, we consider both accuracy of the predicted labels ($\hat{y}$) and performance of rationale prediction in terms of F1, precision, and recall.
We use Pytorch Lightning \citep{falcon2019pytorch} for training with a learning rate of 2e-5 and gradient accumulation over 10 batches for all models. Early stopping was based on validation set loss with a patience of 3, evaluated every fifth of an epoch. Training was performed 
on two 24G NVidia TITAN RTX GPUs.

\subsection{Model Performance}
\label{subsec:p06_results_baseline_comparison}

Table \ref{tab:baseline_results} compares our best overall model against the baselines, and presents the learning strategies used in the models.

\para{Prediction accuracy.} 
For \multirc, this best model includes every proposed modification (sentence-level rationalization, importance embeddings, class weights) except for selective supervision, and yields a 3-point improvement from the baseline accuracy of 68.1\% to 71.2\%. 
We observe a more modest improvement on \fever, with the best model using sentence-level rationalization and importance embeddings, and scoring a 1-point improvement from 90.2\% to 91.5\%. We note, however, that this approaches the accuracy of the model with access to a human rationale oracle (91.6\%).
Finally, we observe a tiny improvement of 0.4\% on \esnli, though our proposed methods do improve upon the baselines of unsupervised and supervised rationale model, which causes a performance drop. 

A McNemar's significance test with 
Bonferroni correction between the best and baseline model finds that the accuracy improvement is significant for \multirc and \fever ($p=$2e-7 and 3e-6 respectively) and not significant for \esnli ($p=0.1$). 
The limited improvement in \esnli echos the performance drop in \figref{fig:combined_unadapted_sufficiency_accuracy} without adaptation, suggesting that human rationales in this dataset are too idiosyncratic to improve model performance.

\begin{table}[]
\small
\centering
\begin{tabular}{@{}llll@{}}
\toprule
\multicolumn{1}{c}{\multirow{2}{*}{\textbf{Method}}}             & \multicolumn{3}{c}{\textbf{Coefficients}}                                                   \\ \cmidrule(l){2-4} 
\multicolumn{1}{c}{}                                             & \multicolumn{1}{c}{\textbf{MultiRC}} & \multicolumn{1}{c}{\textbf{FEVER}} & \textbf{E-SNLI} \\ \midrule
\textbf{Sentences}                                                        & .015***                              & .001                               & -               \\
\textbf{Class weights}                                                    & .017***                              & .007***                            & .005            \\
\textbf{Importance embeddings}                                                       & .012***                              & .006***                            & -.010**         \\
\textbf{Selective supervision} & 0.004                                & -.006***                           & -.032***        \\ \bottomrule
\end{tabular}
\caption{Regression coefficients for effect each proposed method on overall prediction accuracy}
\label{tab:regression_table}
\end{table}

\para{Factor analysis.}
We use regression analysis to understand the impact of the different modifications on model accuracy.
Table \ref{tab:regression_table} suggests that rationale class weighting has the highest positive effect on accuracy across datasets. Importance embeddings have a positive effect for \multirc and \fever and a negative effect for \esnli, while sentence-level rationalization improves only \multirc. 

Selective supervision is found to have a non-existant or negative effect across all three datasets. Table \ref{tab:selective_supervision} details this result, showing model accuracy and rationale performance for the best model with (yes) vs. without (no) selective supervision. 
If our method succeeded, F1 for high-HSA examples would increase from the ``No'' to the ``Yes'' models and remain flat or decrease for low-HSA examples.
Indeed, we observe lower rationale F1 for low-HSA examples, but the rationale F1 also drops substantially for high-HSA examples,
possibly because of the reduced available training data.

\begin{table}[]
\centering
\small
\begin{tabular}{rlrrr}
\hline
\multicolumn{1}{l}{\multirow{2}{*}{\textbf{Dataset}}} & \multirow{2}{*}{\textbf{\begin{tabular}[c]{@{}l@{}}Sel.\\ Sup.\end{tabular}}} & \multicolumn{1}{c}{\multirow{2}{*}{\textbf{Acc.}}} & \multicolumn{2}{c}{\textbf{F1.}} \\ \cline{4-5} 
\multicolumn{1}{l}{}                                  &                                                                               & \multicolumn{1}{c}{}                               & \textbf{High-HSA}   & \textbf{Low-HSA}   \\ \hline
\multirow{2}{*}{\textbf{MultiRC}}                     & \textbf{No}                                                                   & \textbf{71.2}                                      & \textbf{59.3}   & \textbf{57.2}  \\
                                                      & \textbf{Yes}                                                                  & 71.0                                               & 56.2            & 54.1           \\ \hline
\multirow{2}{*}{\textbf{FEVER}}                       & \textbf{No}                                                                   & \textbf{91.5}                                      & \textbf{79.0}   & \textbf{72.5}  \\
                                                      & \textbf{Yes}                                                                  & 90.6                                               & 61.2            & 57.0           \\ \hline
\multirow{2}{*}{\textbf{E-SNLI}}                      & \textbf{No}                                                                   & \textbf{90.1}                                      & \textbf{61.2}   & \textbf{48.0}  \\
                                                      & \textbf{Yes}                                                                  & 88.8                                               & 49.0            & 44.9           \\ \hline
\end{tabular}
\caption{Label accuracy and predicted rationale F1 for high- versus low-HSA examples.}
\label{tab:selective_supervision}
\end{table}

\para{Rationale performance.}
Although our modifications are designed to improve label prediction performance, 
they also improve rationale prediction performance in most cases. The only exception is the reduced precision in \esnli compared to the supervised rationale model.

\begin{table*}[t]
\scriptsize
\centering

\begin{tabular}{@{}p{0.3\linewidth} p{0.3\linewidth} p{0.3\linewidth} @{}}
\toprule
\multicolumn{1}{c}{\textbf{Human rationale}}                                                                                                                                                                                                                                                                                                                                                                                                                                                                                                                                                                                                                                                                                                                                                                                                                                                                                                                                                                                                                                                                                                                                               & \multicolumn{1}{c}{\textbf{Baseline supervised rationale}}                                                                                                                                                                                                                                                                                                                                                                                                                                                                                                                                                                                                                                                                                                                                                                                                                                                                                                                                                                                                                                                                                                                                                                                            & \multicolumn{1}{c}{\textbf{Best model}}                                                                                                                                                                                                                                                                                                                                                                                                                                                                                                                                                                                                                                                                                                                                                                                                                                                                                                                                                                                                                                                                                                                                                                                                                                                                                                                                                                                                                                                                                                                                                                                                                                                                                 \\ \midrule
\multicolumn{3}{c}{\textbf{(A) MultiRC: Best model beats supervised baseline}}                                                                                                                                                                                                                                                                                                                                                                                                                                                                                                                                                                                                                                                                                                                                                                                                                                                                                                                                                                                                                                                                                                                                                                                                                                                                                                                                                                                                                                                                                                                                                                                                                                                                                                                                                                                                                                                                                                                                                                                                                                                                                                                                                                                                                                                                                                                                                                                                                                                                                                                                                                                                                                                                                                                                                                                                                                                                                                                                                                                                                                                                                                                                                                                                                                                                                                                                                                                                                                                                                                                                                                                                                                                                                                                                                                                                                                                                                                                                                                                                                                                                                     \\ \midrule
\definecolor{highlight}{RGB}{250, 215, 160 }\sethlcolor{highlight}\hl{[CLS]} there have been many organisms that have lived in earths past . only a tiny number of them became fossils . still , scientists learn a lot from fossils . fossils are our best clues about the history of life on earth . fossils provide evidence about life on earth . they tell us that life on earth has changed over time . fossils in younger rocks look like animals and plants that are living today . fossils in older rocks are less like living organisms . \hl{fossils}\hl{ }\hl{can}\hl{ }\hl{tell}\hl{ }\hl{us}\hl{ }\hl{about}\hl{ }\hl{where}\hl{ }\hl{the}\hl{ }\hl{organism}\hl{ }\hl{lived}\hl{ }\hl{.}\hl{ }\hl{was}\hl{ }\hl{it}\hl{ }\hl{land}\hl{ }\hl{or}\hl{ }\hl{marine}\hl{ }\hl{?} fossils can even tell us if the water was shallow or deep . fossils can even provide clues to ancient climates . \hl{[SEP]}\hl{ }\hl{what}\hl{ }\hl{can}\hl{ }\hl{we}\hl{ }\hl{tell}\hl{ }\hl{about}\hl{ }\hl{former}\hl{ }\hl{living}\hl{ }\hl{organisms}\hl{ }\hl{from}\hl{ }\hl{fossils}\hl{ }\hl{?}\hl{ }\hl{|}\hl{ }\hl{|}\hl{ }\hl{how}\hl{ }\hl{they}\hl{ }\hl{adapted}\hl{ }\hl{[SEP]} & \definecolor{highlight}{RGB}{230, 176, 170}\sethlcolor{highlight}\hl{[CLS]} there have been many \hl{organisms} that have lived in earths past . only a tiny number \hl{of} them \hl{became} fossils . still , scientists learn a lot from fossils \hl{.}\hl{ }\hl{fossils} are \hl{our} best clues \hl{about} the history of life on earth \hl{.}\hl{ }\hl{fossils}\hl{ }\hl{provide} evidence \hl{about}\hl{ }\hl{life} on earth \hl{.}\hl{ }\hl{they}\hl{ }\hl{tell} us \hl{that}\hl{ }\hl{life} on \hl{earth}\hl{ }\hl{has} changed over time \hl{.} fossils in younger rocks look like animals and plants \hl{that} are living today . fossils in older rocks are less like living organisms \hl{.}\hl{ }\hl{fossils}\hl{ }\hl{can} tell \hl{us}\hl{ }\hl{about}\hl{ }\hl{where} the organism lived . was it land or marine \hl{?} fossils can even tell us if the water was shallow or deep . fossils \hl{can} even provide clues to ancient climates . \hl{[SEP]}\hl{ }\hl{what}\hl{ }\hl{can}\hl{ }\hl{we}\hl{ }\hl{tell}\hl{ }\hl{about}\hl{ }\hl{former}\hl{ }\hl{living}\hl{ }\hl{organisms}\hl{ }\hl{from}\hl{ }\hl{fossils}\hl{ }\hl{?}\hl{ }\hl{|}\hl{ }\hl{|}\hl{ }\hl{how}\hl{ }\hl{they}\hl{ }\hl{adapted}\hl{ }\hl{[SEP]} & \definecolor{highlight}{RGB}{169, 204, 227}\sethlcolor{highlight}\hl{[CLS]} there have been many organisms that have lived in earths past . only a tiny number of them became fossils . still , scientists learn a lot from fossils . \hl{fossils}\hl{ }\hl{are}\hl{ }\hl{our}\hl{ }\hl{best}\hl{ }\hl{clues}\hl{ }\hl{about}\hl{ }\hl{the}\hl{ }\hl{history}\hl{ }\hl{of}\hl{ }\hl{life}\hl{ }\hl{on}\hl{ }\hl{earth}\hl{ }\hl{.}\hl{ }\hl{fossils}\hl{ }\hl{provide}\hl{ }\hl{evidence}\hl{ }\hl{about}\hl{ }\hl{life}\hl{ }\hl{on}\hl{ }\hl{earth}\hl{ }\hl{.}\hl{ }\hl{they}\hl{ }\hl{tell}\hl{ }\hl{us}\hl{ }\hl{that}\hl{ }\hl{life}\hl{ }\hl{on}\hl{ }\hl{earth}\hl{ }\hl{has}\hl{ }\hl{changed}\hl{ }\hl{over}\hl{ }\hl{time}\hl{ }\hl{.}\hl{ }\hl{fossils}\hl{ }\hl{in}\hl{ }\hl{younger}\hl{ }\hl{rocks}\hl{ }\hl{look}\hl{ }\hl{like}\hl{ }\hl{animals}\hl{ }\hl{and}\hl{ }\hl{plants}\hl{ }\hl{that}\hl{ }\hl{are}\hl{ }\hl{living}\hl{ }\hl{today}\hl{ }\hl{.}\hl{ }\hl{fossils}\hl{ }\hl{in}\hl{ }\hl{older}\hl{ }\hl{rocks}\hl{ }\hl{are}\hl{ }\hl{less}\hl{ }\hl{like}\hl{ }\hl{living}\hl{ }\hl{organisms}\hl{ }\hl{.}\hl{ }\hl{fossils}\hl{ }\hl{can}\hl{ }\hl{tell}\hl{ }\hl{us}\hl{ }\hl{about}\hl{ }\hl{where}\hl{ }\hl{the}\hl{ }\hl{organism}\hl{ }\hl{lived}\hl{ }\hl{.} was it land or marine ? fossils can even tell us if the water was shallow or deep . fossils can even provide clues to ancient climates . \hl{[SEP]}\hl{ }\hl{what}\hl{ }\hl{can}\hl{ }\hl{we}\hl{ }\hl{tell}\hl{ }\hl{about}\hl{ }\hl{former}\hl{ }\hl{living}\hl{ }\hl{organisms}\hl{ }\hl{from}\hl{ }\hl{fossils}\hl{ }\hl{?}\hl{ }\hl{|}\hl{ }\hl{|}\hl{ }\hl{how}\hl{ }\hl{they}\hl{ }\hl{adapted}\hl{ }\hl{[SEP]} \\ \midrule
\textbf{Prediction: False}                                                                                                                                                                                                                                                                                                                                                                                                                                                                                                                                                                                                                                                                                                                                                                                                                                                                                                                                                                                                                                                                                                                                                                 & {\textbf{Prediction:  \color[HTML]{FE0000}  True}}                                                                                                                                                                                                                                                                                                                                                                                                                                                                                                                                                                                                                                                                                                                                                                                                                                                                                                                                                                                                                                                                                                                                                                                            & \textbf{Prediction: False}                                                                                                                                                                                                                                                                                                                                                                                                                                                                                                                                                                                                                                                                                                                                                                                                                                                                                                                                                                                                                                                                                                                                                                                                                                                                                                                                                                                                                                                                                                                                                                                                                                                                                              \\ \midrule
\multicolumn{3}{c}{\textbf{(B) FEVER: Human rationale is insufficient}}                                                                                                                                                                                                                                                                                                                                                                                                                                                                                                                                                                                                                                                                                                                                                                                                                                                                                                                                                                                                                                                                                                                                                                                                                                                                                                                                                                                                                                                                                                                                                                                                                                                                                                                                                                                                                                                                                                                                                                                                                                                                                                                                                                                                                                                                                                                                                                                                                                                                                                                                                                                                                                                                                                                                                                                                                                                                                                                                                                                                                                                                                                                                                                                                                                                                                                                                                                                                                                                                                                                                                                                                                                                                                                                                                                                                                                                                                                                                                                                                                                                                                            \\ \midrule
\definecolor{highlight}{RGB}{250, 215, 160 }\sethlcolor{highlight}\hl{[CLS]} legendary entertainment - lrb - also known as legendary pictures or legendary - rrb - is an american media company based in burbank , california . \hl{the}\hl{ }\hl{company}\hl{ }\hl{was}\hl{ }\hl{founded}\hl{ }\hl{by}\hl{ }\hl{thomas}\hl{ }\hl{tu}\hl{ll}\hl{ }\hl{in}\hl{ }\hl{2000}\hl{ }\hl{and}\hl{ }\hl{in}\hl{ }\hl{2005}\hl{ }\hl{,}\hl{ }\hl{concluded}\hl{ }\hl{an}\hl{ }\hl{agreement}\hl{ }\hl{to}\hl{ }\hl{co}\hl{ }\hl{-}\hl{ }\hl{produce}\hl{ }\hl{and}\hl{ }\hl{co}\hl{ }\hl{-}\hl{ }\hl{finance}\hl{ }\hl{films}\hl{ }\hl{with}\hl{ }\hl{warner}\hl{ }\hl{bros}\hl{ }\hl{.}\hl{ }\hl{,}\hl{ }\hl{and}\hl{ }\hl{began}\hl{ }\hl{a}\hl{ }\hl{similar}\hl{ }\hl{arrangement}\hl{ }\hl{with}\hl{ }\hl{universal}\hl{ }\hl{studios}\hl{ }\hl{in}\hl{ }\hl{2014}\hl{ }\hl{.} since 2016 , legendary has been a subsidiary of the chinese conglomerate wanda group . \hl{[SEP]}\hl{ }\hl{legendary}\hl{ }\hl{entertainment}\hl{ }\hl{is}\hl{ }\hl{a}\hl{ }\hl{subsidiary}\hl{ }\hl{of}\hl{ }\hl{warner}\hl{ }\hl{bros}\hl{ }\hl{pictures}\hl{ }\hl{.}\hl{ }\hl{[SEP]}                         & \definecolor{highlight}{RGB}{230, 176, 170}\sethlcolor{highlight}\hl{[CLS]}\hl{ }\hl{legendary} entertainment - lrb - also known as \hl{legendary} pictures or legendary - rr\hl{b} - is an american media \hl{company} based in burbank , california \hl{.} the company \hl{was}\hl{ }\hl{founded} by thomas \hl{tu}ll \hl{in}\hl{ }\hl{2000}\hl{ }\hl{and}\hl{ }\hl{in}\hl{ }\hl{2005}\hl{ }\hl{,}\hl{ }\hl{concluded}\hl{ }\hl{an} agreement \hl{to}\hl{ }\hl{co}\hl{ }\hl{-}\hl{ }\hl{produce}\hl{ }\hl{and} co \hl{-}\hl{ }\hl{finance}\hl{ }\hl{films}\hl{ }\hl{with}\hl{ }\hl{warner} bros \hl{.}\hl{ }\hl{,} and \hl{began}\hl{ }\hl{a} similar \hl{arrangement} with \hl{universal}\hl{ }\hl{studios} in \hl{2014} . \hl{since}\hl{ }\hl{2016}\hl{ }\hl{,}\hl{ }\hl{legendary}\hl{ }\hl{has}\hl{ }\hl{been}\hl{ }\hl{a} subsidiary \hl{of} the \hl{chinese}\hl{ }\hl{conglomerate}\hl{ }\hl{wanda}\hl{ }\hl{group} . \hl{[SEP]}\hl{ }\hl{legendary}\hl{ }\hl{entertainment}\hl{ }\hl{is}\hl{ }\hl{a}\hl{ }\hl{subsidiary}\hl{ }\hl{of}\hl{ }\hl{warner}\hl{ }\hl{bros}\hl{ }\hl{pictures}\hl{ }\hl{.}\hl{ }\hl{[SEP]}                                                                                                              & \definecolor{highlight}{RGB}{169, 204, 227}\sethlcolor{highlight}\hl{[CLS]} legendary entertainment - lrb - also known as legendary pictures or legendary - rrb - is an american media company based in burbank , california . the company was founded by thomas tull in 2000 and in 2005 , concluded an agreement to co - produce and co - finance films with warner bros . , and began a similar arrangement with universal studios in 2014 . \hl{since}\hl{ }\hl{2016}\hl{ }\hl{,}\hl{ }\hl{legendary}\hl{ }\hl{has}\hl{ }\hl{been}\hl{ }\hl{a}\hl{ }\hl{subsidiary}\hl{ }\hl{of}\hl{ }\hl{the}\hl{ }\hl{chinese}\hl{ }\hl{conglomerate}\hl{ }\hl{wanda}\hl{ }\hl{group}\hl{ }\hl{.}\hl{ }\hl{[SEP]}\hl{ }\hl{legendary}\hl{ }\hl{entertainment}\hl{ }\hl{is}\hl{ }\hl{a}\hl{ }\hl{subsidiary}\hl{ }\hl{of}\hl{ }\hl{warner}\hl{ }\hl{bros}\hl{ }\hl{pictures}\hl{ }\hl{.}\hl{ }\hl{[SEP]}                                                                                                                                                                                                                                                                                                                                                                                                                                                                                                                                                                                                                                                                                                                                                                                                                           \\ \midrule
{\textbf{Prediction:  \color[HTML]{FE0000}  Supports}}                                                                                                                                                                                                                                                                                                                                                                                                                                                                                                                                                                                                                                                                                                                                                                                                                                                                                                                                                                                                                                                                                                                                       & {\textbf{Prediction:  \color[HTML]{FE0000}  Supports}}                                                                                                                                                                                                                                                                                                                                                                                                                                                                                                                                                                                                                                                                                                                                                                                                                                                                                                                                                                                                                                                                                                                                                                                        & {\textbf{Prediction:  \color[HTML]{FE0000}  Supports}}                                                                                                                                                                                                                                                                                                                                                                                                                                                                                                                                                                                                                                                                                                                                                                                                                                                                                                                                                                                                                                                                                                                                                                                                                                                                                                                                                                                                                                                                                                                                                                                                                                                                    \\ \midrule
\multicolumn{3}{c}{\textbf{(C) E-SNLI: Supervised baseline beats best model}}                                                                                                                                                                                                                                                                                                                                                                                                                                                                                                                                                                                                                                                                                                                                                                                                                                                                                                                                                                                                                                                                                                                                                                                                                                                                                                                                                                                                                                                                                                                                                                                                                                                                                                                                                                                                                                                                                                                                                                                                                                                                                                                                                                                                                                                                                                                                                                                                                                                                                                                                                                                                                                                                                                                                                                                                                                                                                                                                                                                                                                                                                                                                                                                                                                                                                                                                                                                                                                                                                                                                                                                                                                                                                                                                                                                                                                                                                                                                                                                                                                                                                      \\ \midrule
\definecolor{highlight}{RGB}{250, 215, 160 }\sethlcolor{highlight}\hl{[CLS]} a big dog catches a ball on his nose \hl{[SEP]} a big dog is \hl{sitting}\hl{ }\hl{down} while trying to catch a ball \hl{[SEP]}                                                                                                                                                                                                                                                                                                                                                                                                                                                                                                                                                                                                                                                                                                                                                                                                                                                                                                                                                                              & \definecolor{highlight}{RGB}{230, 176, 170}\sethlcolor{highlight}\hl{[CLS]} a big dog catches a \hl{ball}\hl{ }\hl{on}\hl{ }\hl{his} nose \hl{[SEP]} a big dog is \hl{sitting}\hl{ }\hl{down} while \hl{trying} to catch \hl{a}\hl{ }\hl{ball}\hl{ }\hl{[SEP]}                                                                                                                                                                                                                                                                                                                                                                                                                                                                                                                                                                                                                                                                                                                                                                                                                                                                                                                                                                              & \definecolor{highlight}{RGB}{169, 204, 227}\sethlcolor{highlight}\hl{[CLS]} a big dog \hl{catches}\hl{ }\hl{a} ball on his \hl{nose}\hl{ }\hl{[SEP]} a big \hl{dog} is \hl{sitting}\hl{ }\hl{down}\hl{ }\hl{while}\hl{ }\hl{trying}\hl{ }\hl{to}\hl{ }\hl{catch} a ball \hl{[SEP]}                                                                                                                                                                                                                                                                                                                                                                                                                                                                                                                                                                                                                                                                                                                                                                                                                                                                                                                                                                                                                                                                                                                                                                                                                                                                                                                                                                                                                                      \\ \midrule
\textbf{Prediction: Neutral}                                                                                                                                                                                                                                                                                                                                                                                                                                                                                                                                                                                                                                                                                                                                                                                                                                                                                                                                                                                                                                                                                                                                                               & \textbf{Prediction: Neutral}                                                                                                                                                                                                                                                                                                                                                                                                                                                                                                                                                                                                                                                                                                                                                                                                                                                                                                                                                                                                                                                                                                                                                                                                                & {\textbf{Prediction:  \color[HTML]{FE0000}  Contradiction}}                                                                                                                                                                                                                                                                                                                                                                                                                                                                                                                                                                                                                                                                                                                                                                                                                                                                                                                                                                                                                                                                                                                                                                                                                                                                                                                                                                                                                                                                                                                                                                                                                                                               \\ \bottomrule
\end{tabular}
\caption{Examples of human, supervised baseline, and best model rationales and predictions.}
\label{tab:rationale_examples}
\end{table*}

\begin{table*}[]
\centering
\small
\begin{tabular}{@{}p{0.1\linewidth} p{0.85\linewidth}@{}}
\toprule
\textbf{Source} & \multicolumn{1}{c}{\textbf{Natural language explanation}}                                                                                                                                               \\ \midrule
Human                               & There is no indication that the dog is sitting down while playing catch on his nose.                                                                                                                    \\
Human                               & A dog can catch a ball by not to sitting down.                                                                                                                                                          \\
GPT-3                    & The entailment of this sentence is that the dog is sitting down, and the contradiction would be if the dog was standing up. This sentence is neutral, meaning it doesn't entail or contradict anything. \\ \bottomrule
\end{tabular}
\caption{Examples of natural language explanations for the ``neutral'' prediction on \esnli example from Table \ref{tab:rationale_examples}c. See Appendix \secref{subsec:appendix-gptprompt} for GPT-3 prompt details.}
\label{tab:nle_rationale_examples}
\end{table*}
\subsection{Qualitative Analysis}

Table \ref{tab:rationale_examples} shows three examples, each drawn from a different dataset, to illustrate different outcomes. For each example, we show the human rationale and predicted rationales for both the baseline supervised rationale model and our best overall model. Incorrect predictions are colored red.  

Example \ref{tab:rationale_examples}a shows an instance sampled from \multirc where our best model, with higher recall and sentence-level rationalization, more successfully captures the (sufficient) information present in the human rationale, allowing for a correct prediction where the supervised rationale model fails. 

Example \ref{tab:rationale_examples}b presents a contrasting example from the \fever dataset. The human rationale omits important context, that Legendary Entertainment is a subsidiary of Wanda Group, making it harder to infer that it is \textit{not} a subsidiary of Warner Bros. Our best model succeeds at capturing this snippet in its rationale, but still predicts the incorrect label, illustrating that  
a sufficient (for humans) rationale does not always produce a correct label.

Finally, example \ref{tab:rationale_examples}c shows a case where the baseline supervised rationale model succeeds while our best model fails. This is a hard-to-interpret example, mainly a demonstration of the limitations of rationales as an explanatory device for certain kinds of task. This begs a question: how relevant are rationales as an explanation or learning mechanism when models like GPT-3 \cite{brown_language_2020} are increasingly capable of human-level \textit{natural language} explanations (Table \ref{tab:nle_rationale_examples})?

Our position is that however an explanation is presented, meaning is still localized within text, so rationales can still serve as a useful interface for scrutinizing or controlling model logic, even if they require additional translation to be comprehensible to humans. Works that hybridize the two ideas such as \citet{zhao_lirex_2020} may represent a good way of resolving this issue.

\section{Discussion}

The analysis in section \secref{sec:analysis} explores the limits of potential improvement from \learningfromexplanation. 
It suggests two insights toward improved \learningfromexplanation: 1) 
that insofar as they boost model accuracy, 
not all human rationale tokens are equally valuable, e.g., with false positives causing less degradation than false negatives; 
and 2) we could in principle boost label accuracy with good rationale accuracy on useful (high-SA) rationales and low accuracy on useless (low-SA) ones. 

We exploit these two insights with four modifications to the baseline architecture. Three of these diverge from flat rationale supervision accuracy: rationale supervision class weighting, sentence-level rationalization, and importance embeddings. The last, selective supervision, pursues utility-discriminative weighting during model training. 

Taken together, our proposed methods yield a substantial 3\% improvement over baseline performance for \multirc, a 1\% improvement on \fever, and a tiny .4\% improvement on \esnli, mirroring the potential improvements observed in the analysis. We find that all three token supervision methods are useful in achieving this, while selective supervision has a marginal or negative effect.

In summary, our results support the potential for \learningfromexplanation in certain datasets, and demonstrate the importance of understanding the properties of human rationales to properly exploit them for this purpose.
We believe that these two insights are useful steps towards effective \learningfromexplanation, and could yield even greater improvements if operationalized optimally.

\para{Limitation.}
A limitation of our analysis is that all three datasets are document-query style reading comprehension tasks, as opposed to, e.g., sentiment analysis. Because of the popularity of this type of task in NLP benchmarks, this type of dataset represents a majority of what is available in 
the ERASER collection \cite{deyoung_eraser_2019}. By contrast, sentiment is often scattered throughout a text, so human rationales for sentiment are likely to contain redundant signal, which could impact their predictive utility. We leave a more comprehensive survey of NLP tasks for future work.

\para{Acknowledgments.}
We thank anonymous reviewers for their feedback, and members of the Chicago Human+AI Lab for their insightful suggestions. 
This work is supported in part by research awards from Amazon, IBM, Salesforce, and NSF IIS-2126602.

\bibliography{anthology,custom,sam_zotero}
\bibliographystyle{acl_natbib}

\newpage 
\appendix

\begin{table*}[]
\small
\centering
\begin{tabular}{@{}rlrrrrrr@{}}
\toprule
\multicolumn{1}{l}{\multirow{2}{*}{\textbf{Dataset}}} & \multirow{2}{*}{\textbf{Method}} & \multicolumn{1}{c}{\multirow{2}{*}{\textbf{Role}}} & \multicolumn{1}{c}{\multirow{2}{*}{\textbf{Accuracy}}} & \multicolumn{3}{c}{\textbf{Rationale prediction}}                                                              & \multicolumn{1}{c}{\multirow{2}{*}{\textbf{\begin{tabular}[c]{@{}c@{}}Human \\ Suff. Acc.\end{tabular}}}} \\ \cmidrule(lr){5-7}
\multicolumn{1}{l}{}                                  &                                  & \multicolumn{1}{c}{}                               & \multicolumn{1}{c}{}                                   & \multicolumn{1}{c}{\textbf{F1}} & \multicolumn{1}{c}{\textbf{Precision}} & \multicolumn{1}{c}{\textbf{Recall}} & \multicolumn{1}{c}{}                                                                                      \\ \midrule
\multirow{8}{*}{\textbf{MultiRC}}                     & \textbf{Sentences}               & Best with                                          & 71.2                                                   & 57.1                            & 44.9                                   & 78.4                                & 74.5                                                                                                      \\
                                                      & \textbf{Sentences}               & Best without                                       & 70.6                                                   & 41.6                            & 27.7                                   & 84.1                                & 75.8                                                                                                      \\ \cmidrule(l){2-8} 
                                                      & \textbf{Class-weights}           & Best with                                          & 71.2                                                   & 57.1                            & 44.9                                   & 78.4                                & 74.5                                                                                                      \\
                                                      & \textbf{Class-weights}           & Best without                                       & 70.8                                                   & 55.2                            & 66.1                                   & 47.4                                & 76.5                                                                                                      \\ \cmidrule(l){2-8} 
                                                      & \textbf{Importance embeddings}   & Best with                                          & 71.2                                                   & 57.1                            & 44.9                                   & 78.4                                & 74.5                                                                                                      \\
                                                      & \textbf{Importance embeddings}   & Best without                                       & 71.0                                                   & 53.6                            & 39.7                                   & 82.5                                & 75.8                                                                                                      \\ \cmidrule(l){2-8} 
                                                      & \textbf{Selective supervision}   & Best with                                          & 71.0                                                   & 53.6                            & 39.7                                   & 82.5                                & 75.8                                                                                                      \\
                                                      & \textbf{Selective supervision}   & Best without                                       & 71.2                                                   & 57.1                            & 44.9                                   & 78.4                                & 74.5                                                                                                      \\ \midrule
\multirow{8}{*}{\textbf{FEVER}}                       & \textbf{Sentences}               & Best with                                          & 91.5                                                   & 81.2                            & 83.5                                   & 79.1                                & 91.6                                                                                                      \\
                                                      & \textbf{Sentences}               & Best without                                       & 91.3                                                   & 72.4                            & 61.3                                   & 88.5                                & 91.6                                                                                                      \\ \cmidrule(l){2-8} 
                                                      & \textbf{Class-weights}           & Best with                                          & 91.5                                                   & 79.6                            & 73.1                                   & 87.3                                & 91.8                                                                                                      \\
                                                      & \textbf{Class-weights}           & Best without                                       & 91.5                                                   & 81.2                            & 83.5                                   & 79.1                                & 91.6                                                                                                      \\ \cmidrule(l){2-8} 
                                                      & \textbf{Importance embeddings}   & Best with                                          & 91.5                                                   & 81.2                            & 83.5                                   & 79.1                                & 91.6                                                                                                      \\
                                                      & \textbf{Importance embeddings}   & Best without                                       & 91.4                                                   & 80.0                            & 74.9                                   & 85.9                                & 91.8                                                                                                      \\ \cmidrule(l){2-8} 
                                                      & \textbf{Selective supervision}   & Best with                                          & 90.6                                                   & 56.4                            & 41.4                                   & 88.6                                & 90.4                                                                                                      \\
                                                      & \textbf{Selective supervision}   & Best without                                       & 91.5                                                   & 81.2                            & 83.5                                   & 79.1                                & 91.6                                                                                                      \\ \midrule
\multirow{6}{*}{\textbf{E-SNLI}}                      & \textbf{Class-weights}           & Best with                                          & 90.1                                                   & 59.6                            & 45.5                                   & 86.2                                & 92.3                                                                                                      \\
                                                      & \textbf{Class-weights}           & Best without                                       & 89.9                                                   & 62.2                            & 55.7                                   & 70.4                                & 92.0                                                                                                      \\ \cmidrule(l){2-8} 
                                                      & \textbf{Importance embeddings}   & Best with                                          & 90.1                                                   & 59.6                            & 45.5                                   & 86.2                                & 92.3                                                                                                      \\
                                                      & \textbf{Importance embeddings}   & Best without                                       & 89.9                                                   & 33.5                            & 20.2                                   & 100.0                               & 72.5                                                                                                      \\ \cmidrule(l){2-8} 
                                                      & \textbf{Selective supervision}   & Best with                                          & 88.8                                                   & 49.0                            & 33.2                                   & 93.4                                & 84.0                                                                                                      \\
                                                      & \textbf{Selective supervision}   & Best without                                       & 90.1                                                   & 59.6                            & 45.5                                   & 86.2                                & 92.3                                                                                                      \\ \bottomrule
\end{tabular}
\caption{Comparison of best model with each proposed factor against best model without that factor. }
\label{tab:factor_comparison}
\end{table*}

\section{Detailed Factor Analysis}

Table \ref{tab:factor_comparison} compares, for each proposed method, the performance of the best model using that method and the best model not using it. The story shown here is similar to the regression analysis in Table \ref{tab:regression_table}, but one new insight is that the improvement in model prediction performance appears to be driven by the sentence-level rationalization method, as it cuts down on stray tokens dropped from or added to the predicted rationales. 

\section{Rationale Perturbation on FEVER and E-SNLI}
\label{subsec:appendix-generalization}
Furthering the analysis in \secref{subsec:rationale-recall-analysis}, we extend the human rationale perturbation experiment to FEVER and E-SNLI. 

\begin{figure*}[h]
\centering
\begin{subfigure}[t]{0.41\textwidth}
    \centering
    \includegraphics[width=0.9\textwidth]{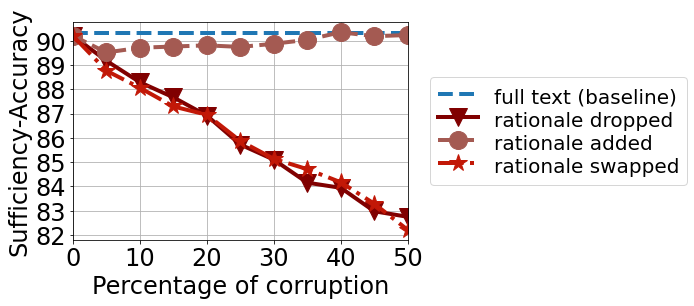}
    \caption{All samples}
    \label{fig:fever_perturbation_plots_all}
\end{subfigure}
\begin{subfigure}[t]{0.28\textwidth}
    \centering
    \includegraphics[width=0.85\textwidth, trim=0 0 270 0, clip]{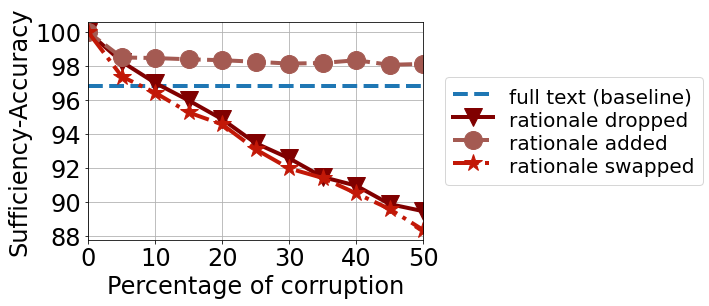}
    \caption{Human sufficiency-accuracy = 1}
    \label{fig:fever_perturbation_plots_suffacc_1}
\end{subfigure}
\begin{subfigure}[t]{0.28\textwidth}
    \centering
    \includegraphics[width=0.83\textwidth, trim=0 0 270 0, clip]{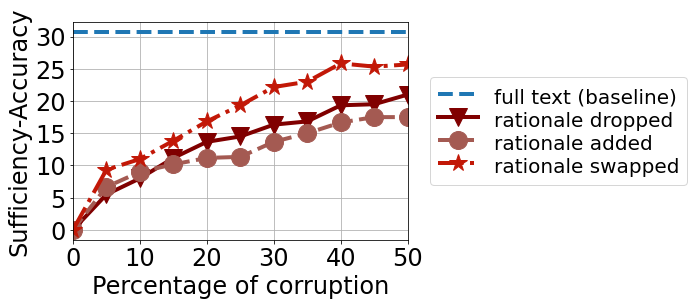}
    \caption{Human sufficiency-accuracy = 0}
    \label{fig:fever_perturbation_plots_suffacc_0}
\end{subfigure}
\caption{Performance of corrupted rationale for FEVER. Model performance drops below baseline accuracy immediately on both dropping human rationales (i.e., recall $\downarrow$) and adding non-rationale tokens (i.e., precision $\downarrow$). For HSA $=1$, model performance remains consistently above baseline on adding non-rationale tokens (i.e. precision $\downarrow$)}
\label{fig:fever_perturbation_plots}
\end{figure*}

\figref{fig:fever_perturbation_plots} show the result for \fever. \figref{fig:fever_perturbation_plots_all} shows that the baseline accuracy is so high for this dataset that to match just the baseline accuracy for FEVER, we require near perfect prediction of human rationales. 

Moreover, even for documents with HSA $=1$, the model performance drops below baseline on dropping just $\sim$ 10\% tokens (synonymous with rationale recall = $\sim$0.9) in \figref{fig:fever_perturbation_plots_suffacc_1}. Interestingly, the model performance remains consistently above the baseline when adding non-rationale tokens (synonymous with decreasing rationale precision). In comparison, the model performance for MultiRC in \figref{fig:multirc_perturbation_plots_suffacc_1} drops below baseline after dropping $\sim$50\% of the tokens. 

For \fever examples with HSA $=0$ (\figref{fig:fever_perturbation_plots_suffacc_0}), the model performance remains below the baseline accuracy consistently, supporting the second hypothesis in \secref{subsec:rationale-recall-analysis}. The near-perfect need to predict rationales in FEVER may explain behind the difference in improvements of model performance between MultiRC and FEVER.

\begin{figure*}[h]
\centering
\begin{subfigure}[t]{0.41\textwidth}
    \centering
    \includegraphics[width=0.9\textwidth]{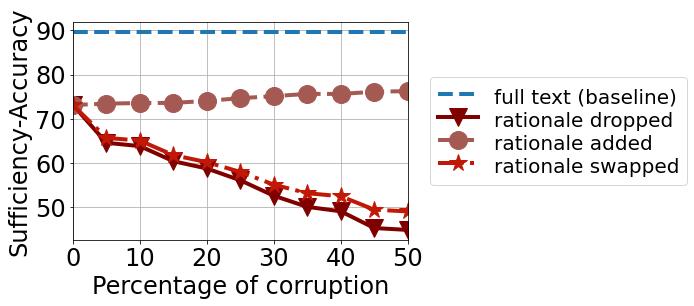}
    \caption{All samples}
    \label{fig:ensli_perturbation_plots_all}
\end{subfigure}
\begin{subfigure}[t]{0.28\textwidth}
    \centering
    \includegraphics[width=0.85\textwidth, trim=0 0 270 0, clip]{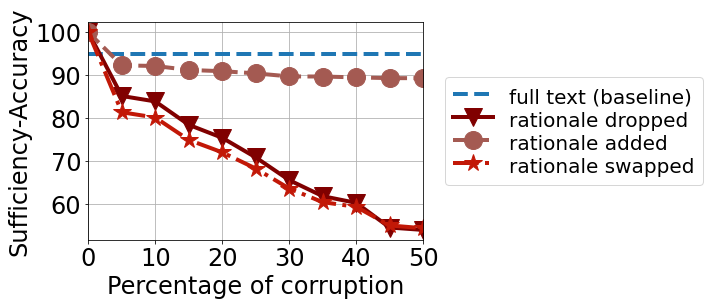}
    \caption{Human sufficiency-accuracy = 1}
    \label{fig:ensli_perturbation_plots_suffacc_1}
\end{subfigure}
\begin{subfigure}[t]{0.28\textwidth}
    \centering
    \includegraphics[width=0.83\textwidth, trim=0 0 270 0, clip]{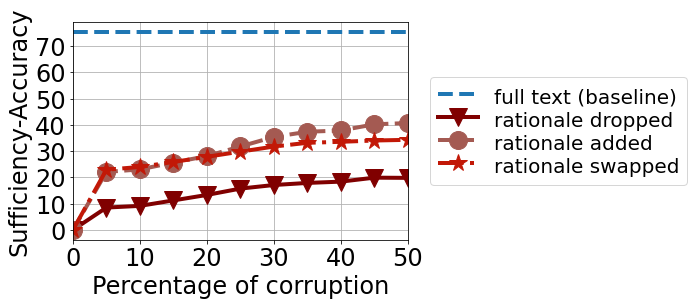}
    \caption{Human sufficiency-accuracy = 0}
    \label{fig:ensli_perturbation_plots_suffacc_0}
\end{subfigure}
\caption{Performance of corrupted rationales for E-SNLI. Model performance for human rationale remains below baseline accuracy and slightly increases with addition of non-rationale tokens (i.e. precision $\downarrow$).  Even for HSA $=1$, model performance drops below baseline accuracy at just $\sim$4\% corruption.}
\label{fig:ensli_perturbation_plots}
\end{figure*}

\figref{fig:ensli_perturbation_plots} covers \esnli. We see that the model performance decreases after dropping rationale tokens (signifying decreasing recall) and it consistently remains below the baseline. In contrast, the model performance shows a slight improvement after adding non-rationale tokens (signifying decrease in rationale precision). Moreover, for documents with HSA $=1$, the model performance drops below baseline at $\sim$3\% for dropping and swapping rationale tokens, where as the model performance plateaus with addition of non-rationale tokens. These insights highlights the substantial challenges in learning from explanations for E-SNLI.


\def \dataset {fever}
\def \basedir {new_plots}

\def \dataset {ensli}
\def \basedir {new_plots}

\section{Rationale Perturbation for Adapted Models}

We perform the same perturbation analysis on calibrated model trained on both full and rationalized input, for which distribution shift from masking are less of a concern. 

\begin{figure*}[h]
\centering
\begin{subfigure}[t]{0.41\textwidth}
    \centering
    \includegraphics[width=0.9\textwidth]{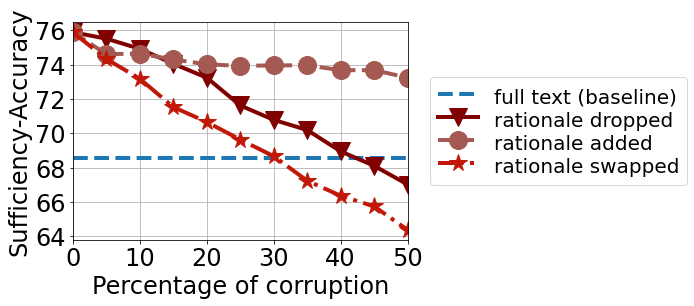}
    \caption{All samples}
    \label{fig:multirc_cal_perturbation_plots_all}
\end{subfigure}
\begin{subfigure}[t]{0.28\textwidth}
    \centering
    \includegraphics[width=0.85\textwidth, trim=0 0 270 0, clip]{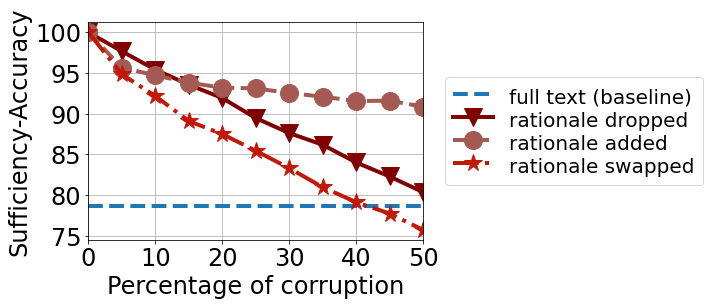}
    \caption{Human sufficiency-accuracy = 1}
    \label{fig:multirc_cal_perturbation_plots_suffacc_1}
\end{subfigure}
\begin{subfigure}[t]{0.28\textwidth}
    \centering
    \includegraphics[width=0.83\textwidth, trim=0 0 270 0, clip]{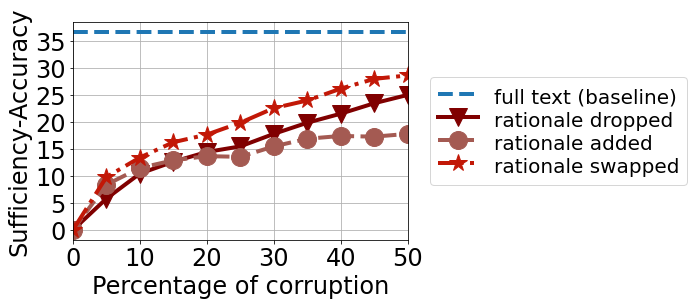}
    \caption{Human sufficiency-accuracy = 0}
    \label{fig:multirc_cal_perturbation_plots_suffacc_0}
\end{subfigure}
\caption{Performance of corrupted rationales for MultiRC using a calibrated model.
Model performance decreases consistently when we drop human rationales (i.e., recall $\downarrow$), where as the model performance stays high as we add non-rationale tokens (i.e., precision $\downarrow$).
The impact of recall is moderated when HSA$=1$.}
\label{fig:multirc_cal_perturbation_plots}
\end{figure*}

\begin{figure*}[h]
\centering
\begin{subfigure}[t]{0.41\textwidth}
    \centering
    \includegraphics[width=0.9\textwidth]{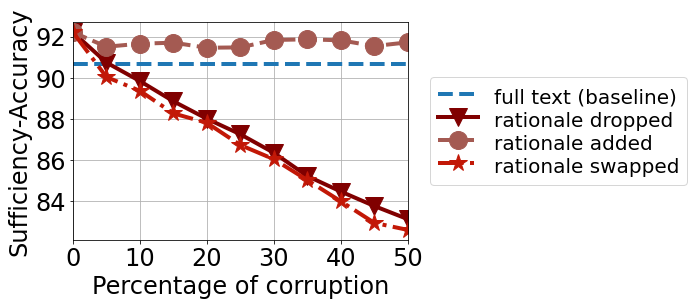}
    \caption{All samples}
    \label{fig:fever_cal_perturbation_plots_all}
\end{subfigure}
\begin{subfigure}[t]{0.28\textwidth}
    \centering
    \includegraphics[width=0.85\textwidth, trim=0 0 270 0, clip]{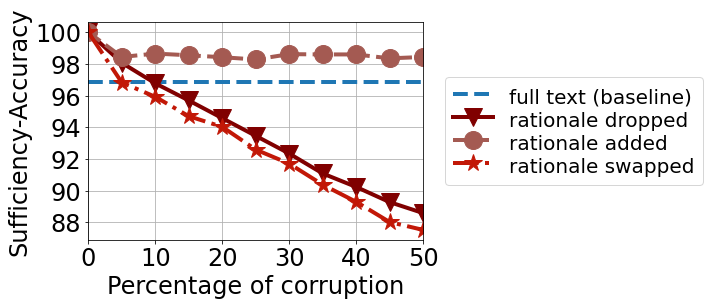}
    \caption{Human sufficiency-accuracy = 1}
    \label{fig:fever_cal_perturbation_plots_suffacc_1}
\end{subfigure}
\begin{subfigure}[t]{0.28\textwidth}
    \centering
    \includegraphics[width=0.83\textwidth, trim=0 0 270 0, clip]{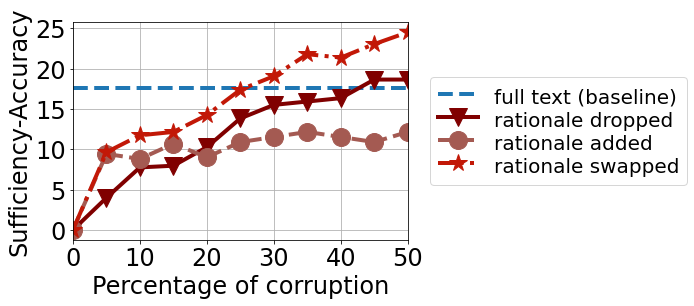}
    \caption{Human sufficiency-accuracy = 0}
    \label{fig:fever_cal_perturbation_plots_suffacc_0}
\end{subfigure}
\caption{Performance of corrupted rationales for FEVER using a calibrated model.
Model performance decreases quickly when we drop human rationales (i.e., recall $\downarrow$), where as the model performance remains above baseline as we add non-rationale tokens (i.e., precision $\downarrow$).}
\label{fig:fever_cal_perturbation_plots}
\end{figure*}

In \figref{fig:multirc_cal_perturbation_plots}, for MultiRC, we find that model performance plateaus with addition of non-rationale tokens and drops quickly with rationale tokens even for a calibrated model. This observation is consistent for FEVER (\figref{fig:fever_cal_perturbation_plots}). 

For \esnli, we find different properties using a calibrated BERT model compared to the standard BERT model show in \figref{fig:ensli_perturbation_plots_all}. 

In contrast to MultiRC and FEVER, we find that the model performance drops more rapidly with the addition of non-rationale tokens compared to removal of rationale tokens. This is consistent for documents with HSA $=1$, suggesting that for E-SNLI, rationale precision maybe more important when using a calibrated model. Similar to FEVER, we see the model performance drop below the baseline with very little corruption of rationales, echoing the need to perfectly mimic human rationalization for effective learning from rationales for this dataset.

\label{subsubsec:calibrated-model}

\def \dataset {multirc-cal}
\def \basedir {new_plots}

\def \dataset {fever-cal}
\def \basedir {new_plots}

\def \dataset {ensli-cal}
\def \basedir {new_plots}
\begin{figure*}[h]
\centering
\begin{subfigure}[t]{0.41\textwidth}
    \centering
    \includegraphics[width=0.9\textwidth]{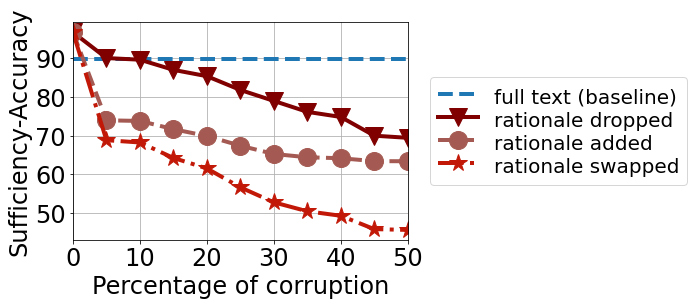}
    \caption{All samples}
    \label{fig:ensli_cal_perturbation_plots_all}
\end{subfigure}
\begin{subfigure}[t]{0.28\textwidth}
    \centering
    \includegraphics[width=0.85\textwidth, trim=0 0 270 0, clip]{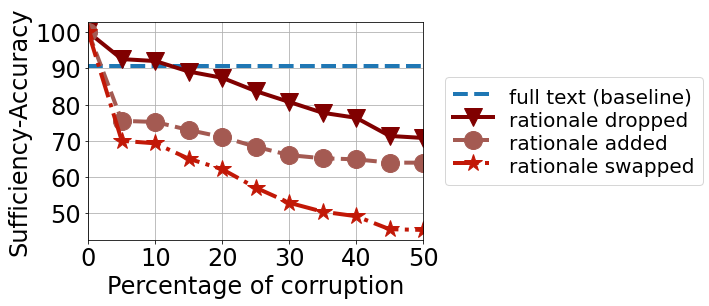}
    \caption{Human sufficiency-accuracy = 1}
    \label{fig:ensli_cal_perturbation_plots_suffacc_1}
\end{subfigure}
\begin{subfigure}[t]{0.28\textwidth}
    \centering
    \includegraphics[width=0.83\textwidth, trim=0 0 270 0, clip]{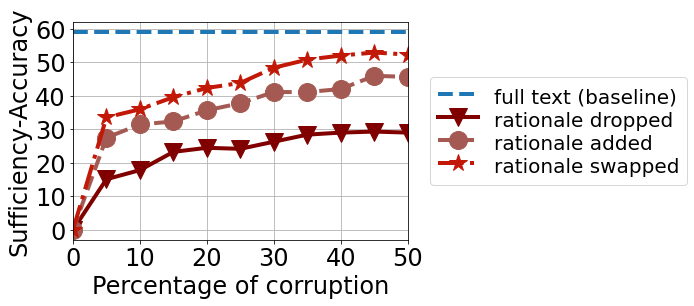}
    \caption{Human sufficiency-accuracy = 0}
    \label{fig:ensli_cal_perturbation_plots_suffacc_0}
\end{subfigure}
\caption{Performance of corrupted rationales for E-SNLI using a calibrated model.
Model performance decreases quickly when we add non- rationale tokens (i.e., precision $\downarrow$), where as the model performance drops less rapidly as we drop rationale tokens (i.e., recall $\downarrow$).
}
\label{fig:ensli_cal_perturbation_plots}
\end{figure*}

\section{GPT-3 Prompt}
\label{subsec:appendix-gptprompt}
We generate a zero-shot GPT-3 \cite{brown_language_2020} explanation using the Davinci model variant on the OpenAI playground\footnote{https://beta.openai.com/playground}, and a modified version of the prompt proposed by \citet{wiegreffe_reframing_2021}: 

\begin{quote}

Let's explain classification decisions. 

A big dog catches a ball on his nose. 

question: A big dog is sitting down while trying to catch a ball. 

entailment, contradiction, or neutral? 

\end{quote}

A second step prompting for an explanation is not needed, as GPT-3 gives its prediction in the form of a natural language explanation. 

\end{document}